\documentclass[conference]{IEEEtran}
\IEEEoverridecommandlockouts
\usepackage{cite}
\usepackage{amsmath,amssymb,amsfonts}
\usepackage{algorithmic}
\usepackage{graphicx}
\usepackage{textcomp}
\usepackage[table,xcdraw]{xcolor}
\usepackage{tikz}

\AtBeginDocument{%
 \providecommand\BibTeX{{%
 \normalfont B\kern-0.5em{\scshape i\kern-0.25em b}\kern-0.8em\TeX}}}
 
\newcommand\copyrighttext{%
  \footnotesize \textcopyright \the\year{} IEEE. Personal use of this material is permitted.  Permission from IEEE must be obtained for all other uses, in any current or future media, including reprinting/republishing this material for advertising or promotional purposes, creating new collective works, for resale or redistribution to servers or lists, or reuse of any copyrighted component of this work in other works.}

\newcommand\copyrightnotice{%
\begin{tikzpicture}[remember picture,overlay]
\node[anchor=south,yshift=10pt] at (current page.south) {\fbox{\parbox{\dimexpr0.75\textwidth-\fboxsep-\fboxrule\relax}{\copyrighttext}}};
\end{tikzpicture}%
}

\begin{document}
% More Than Meets The Eye: Leveraging Transformers for Advanced Named Entity Recognition
\title{Evaluating the Efficacy of AI Techniques in Textual Anonymization: A Comparative Study
}

\author{Dimitris Asimopoulos\IEEEauthorrefmark{2}, Ilias Siniosoglou\IEEEauthorrefmark{1}\IEEEauthorrefmark{2}, Vasileios Argyriou\IEEEauthorrefmark{3}, Sotirios K. Goudos\IEEEauthorrefmark{4}, Konstantinos E. Psannis\IEEEauthorrefmark{5}, \\Nikoleta Karditsioti\IEEEauthorrefmark{6}, Theocharis Saoulidis\IEEEauthorrefmark{6} and Panagiotis Sarigiannidis\IEEEauthorrefmark{1}\IEEEauthorrefmark{2}

\thanks{\IEEEauthorrefmark{1} I. Siniosoglou and P. Sarigiannidis are with the Department of Electrical and Computer Engineering, University of Western Macedonia, Kozani, Greece - \texttt{E-Mail: \{isiniosoglou, psarigiannidis\}@uowm.gr}}

\thanks{\IEEEauthorrefmark{2} I. Siniosoglou, D. Asimopoulos and P. Sarigiannidis are with the R\&D Department, MetaMind Innovations P.C., Kozani, Greece - \texttt{E-Mail: \{isiniosoglou, dasimopoulos, psarigiannidis\}@metamind.gr}}

\thanks{\IEEEauthorrefmark{3} V. Argyriou is with the Department of Networks and Digital Media, Kingston University, Kingston upon Thames, United Kingdom - \texttt{E-Mail: vasileios.argyriou@kingston.ac.uk}}

\thanks{\IEEEauthorrefmark{4} S. K. Goudos is with the Physics Department, Aristotle University of Thessaloniki, Thessaloniki, Greece   - \texttt{E-Mail: sgoudo@physics.auth.gr}}

\thanks{\IEEEauthorrefmark{5} K. E. Psannis is with the Department of Applied Informatics, School of Information Sciences, University of Macedonia, Thessaloniki, Greece   - \texttt{E-Mail: kpsannis@uom.edu.gr}}

\thanks{\IEEEauthorrefmark{6}  N. Karditsioti and T. Saoulidis are with Sidroco Holdings Ltd., Nicosia, Cyprus - \texttt{E-Mail: \{nkarditsioti, hsaoulidis\}@sidroco.com}}
}

\maketitle
\copyrightnotice

\begin{abstract}
In the digital era, with escalating privacy concerns, it's imperative to devise robust strategies that protect private data while maintaining the intrinsic value of textual information. This research embarks on a comprehensive examination of text anonymisation methods, focusing on Conditional Random Fields (CRF), Long Short-Term Memory (LSTM), Embeddings from Language Models (ELMo), and the transformative capabilities of the Transformers architecture. Each model presents unique strengths since LSTM is modeling long-term dependencies, CRF captures dependencies among word sequences, ELMo delivers contextual word representations using deep bidirectional language models and Transformers introduce self-attention mechanisms that provide enhanced scalability. Our study is positioned as a comparative analysis of these models, emphasising their synergistic potential in addressing text anonymisation challenges. Preliminary results indicate that CRF, LSTM, and ELMo individually outperform traditional methods. The inclusion of Transformers, when compared alongside with the other models, offers a broader perspective on achieving optimal text anonymisation in contemporary settings.
\end{abstract}

\begin{IEEEkeywords}
Data anonymisation, text anonymisation, LSTM, CRF, ELMo, Transformers, B2G
\end{IEEEkeywords}

\section{Introduction}
\label{Introduction}
In the contemporary digital age, the proliferation of data-intensive applications and services has fostered a landscape where the collection and analysis of massive datasets have become commonplace. This is especially the case in rapid and cross-domain applications, leveraging high fidelity architecutres, like Open-RAN. This wealth of data, often encompassing sensitive or personally identifiable information (PII), necessitates robust mechanisms for protecting individual privacy while retaining the utility of the data for analytical processes \cite{payne2020privacy} in B2G networks. This in enhanced by the forthcoming data streamlining mechanisms that advanced 6G functionalities promise. Textual and numerical anonymisation \cite{Lison2021} emerges as a cardinal strategy in this endeavour, facilitating the obfuscation of sensitive elements within datasets to thwart potential misuse.
Anonymisation techniques have historically employed a range of strategies, spanning from simple methods such as data masking and perturbation to more sophisticated approaches leveraging advanced machine learning algorithms. Recently, dl approaches \cite{mase2023facial} have demonstrated a promising avenue for advancing the novelty in data anonymisation, extending the range techniques to include methods grounded in natural language processing and complex numerical transformations.

In this study, we explore advanced ML techniques for data anonymisation, focusing on CRF, LSTM, ELMo, and Transformer models. CRFs offer a probabilistic framework for labelling and segmenting sequential data, a property integral in textual anonymisation. On the other hand, LSTMs facilitate the modelling of sequences and their long dependencies \cite{Siniosoglou2023}, providing a rich ground for both textual and numerical anonymisation and ELMo, a deep contextualized word representation.

%Our primary contribution centres on the meticulous assessment of a transformer model fine-tuned to meet the specific needs of our study in the context of Named Entity Recognition (NER). This approach is leveraged to perform a comparative analysis with established models such as CRF, LSTM, and ELMo, which have previously demonstrated efficacy in various NLP tasks\cite{goyal2023survey}.
%By evaluating the performance across the  NER dataset our study navigates through the unique characteristics and challenges presented, providing a fertile ground for a comprehensive evaluation. This initiative not only offers a fresh perspective on the capabilities of transformer models but also enriches the existing body of research by introducing a solution that aspires to optimise performance in name entity recognition tasks.
In the context of Name Entity Recognition (NER), our main contribution is the evaluation of a fine-tuned transformer model in order to meet the requirements of our study. This approach is used to compare the model with well-known models such us CRF, LSTM and Elmo, which have been shown to be effective in a number of NLP tasks in the past\cite{goyal2023survey}. Our study provides a fertile ground for a comprehensive evaluation by navigating through the distinctive qualities and obstacles given by analyzing the performance across the NER dataset. This study contributes to the existing body of research and also provides a new point of view on the transformer models, by introducing a solution that aims to optimize the performance in NER tasks. \cite{friebely2022analyzing}.

The remainder of this paper is organised as follows: Section \ref{Related Work} presents the related work. Section \ref{Methodology} delves into the proposed methodology and Section \ref{Experimental Results} presents a comparative study of results. Finally, Section \ref{Conclusion} concludes this work.

\section{Related Work}
\label{Related Work}

%In the pursuit of managing sensitive data, text anonymisation and privacy protection have garnered pivotal attention across various domains such as healthcare and education. Several research endeavours \cite{Asimopoulos2023}, have underscored the efficacy of deep learning models in addressing these challenges. One notable contribution in this domain is by Rosario Catelli et al.\cite{catelli2021combining}, who proposed a novel method that amalgamates contextualized word representations with a Bi-LSTM+CRF architecture, primarily for clinical de-identification. This innovative approach, through its adept understanding of clinical context, facilitates the accurate recognition and labelling of sensitive information spans, thereby enhancing de-identification performance. 
Data privacy and text anonymisation are crucial for protecting sensitive data in different domains, including healthcare and finance. Numerous studies have presented the effectiveness of deep learning models in providing solutions in this kind of problem \cite{Asimopoulos2023}. In \cite{catelli2021combining} provide a notable work in this field where they propose a novel method that combines contextualized word representations with a Bi-LSTM+CRF architecture. This method improves de-identification performance, through its ability to understand clinical environments, making easier the identification and classification of sensitive information. 

%A distinct avenue of research has been explored by Cillian Berragan et al.\cite{berragan2023transformer}, who developed and evaluated custom-built NER models specifically for place name extraction from online text. Their research demonstrated a significant advancement in performance, with their best model achieving an F1 score of 0.93, showcasing the potential of such models in extracting geographic information from volunteered online sources. Addressing the challenges faced by low-resourced languages, Ridewaan Hanslo \cite{hanslo2022deep} delved into the evaluation of Deep Learning transformer architecture models for NER. The study revealed a pronounced improvement in the performance of transformer models when fine-tuned, surpassing other conventional Neural Network and Machine Learning models, including CRF. In a similar vein, Saadullah Amin et al.\cite{amin2021t2ner} presented T2NER, a Transformers-based Transfer Learning framework for NER, highlighting the recent shifts from traditional LSTM networks to advanced deep transformer models. This framework, developed in PyTorch, supports a myriad of transfer learning scenarios, offering a unified and extensible platform for both research and real-world applications in NER.
\cite{berragan2023transformer} have investigated a different approach by creating and assessing NER models that are specially designed for place name extraction from online text.Their best model obtained an F1 score of 0.93, indicating the possibility of such models in extracting geographic information from open sources. Their research showed a notable improvement in performance. Addressing the issues presented by low-resourced languages, \cite{hanslo2022deep} delved into the evaluation of Deep Learning transformer architecture models for NER. The results showed that by fine-tuning the transformers models their performance is much better that the traditional Neural Network and Machine Learning models, such as CRF. Similarly, \cite{amin2021t2ner} introduced T2NER, a Transformers-based Transfer Learning framework for NER, emphasizing the recent transition from sophisticated deep transformer models to regular LSTM networks. Using PyTorch, this framework provides an unified and flexible platform for study and application in NER, covering in this way a wide range of transfer learning scenarios.

Furthermore, the significance of predictive analytics in education has been emphasized by \cite{uliyan2021deep}, where a model based on Bidirectional Long Short-Term Memory (BLSTM) and CRF showcased its proficiency in predicting student retention, thereby aiding institutions in strategic decision-making. In addition, \cite{pilan2022text} addressed the scarcity of standardized evaluation procedures for text anonymisation by introducing a specific corpus and assessment framework. This initiative facilitates the comparison and advancement of current methods through a defined benchmark, the TAB.

% This paper compares text anonymisation strategies and models, focusing on CRF, LSTM, ELMo, and Transformers. It evaluates their strengths and capabilities, aiming to identify their distinct attributes and efficacy, contributing to efforts to ensure sensitive information privacy while maintaining data utility.

% In light of the aforementioned studies, this paper aims to advance the discourse in text anonymisation by conducting a comprehensive comparison of cutting-edge strategies and models. Our work meticulously evaluates and contrasts the strengths and capabilities of CRF, LSTM, ELMo, and Transformers. This comparative approach seeks to unveil the distinct attributes and efficacies of each model, thereby contributing to the ongoing efforts to ensure the privacy of sensitive information while maintaining the utility of the processed data.

\section{Methodology}
\label{Methodology}

\subsection{Data Anonymisation}
\label{Data Anonymisation}

% Anonymisation remains a pivotal process in safeguarding sensitive data, involving the meticulous identification and subsequent neutralization of such information in documents. This intricate procedure unfolds in two fundamental stages: the recognition of sensitive elements, followed by their elimination or alteration. 

Anonymisation is an essential procedure for safeguarding sensitive data, which entails the identification and elimination of such information in documents through two fundamental phases.A significant number of techniques have been developed in the field of privacy to correctly identify sensitive data. The first step of each technique based on NER principles \cite{baigang2023review} is the independent evaluation of referent entities to detect organisational, personal and more potential private information. After this step, a strategy mapped to the data structure and organisations relevant to cybersecurity, is implemented. 

These strategies for neutralization include the following:
\begin{itemize}
\item \textbf{Removal}: where references to private information are replaced with placeholder characters, effectively erasing the sensitive data.
\item \textbf{Categorization}: which involves substituting labels for references, thereby indicating the general type or category of the erased information without divulging specifics.
\item \textbf{Pseudonymisation}: a process of replacing sensitive records with alternate versions retaining the same data category, applicable in specialized data contexts.
\end{itemize}

\begin{figure}[httb]
\centering
\includegraphics[width=0.7\linewidth]{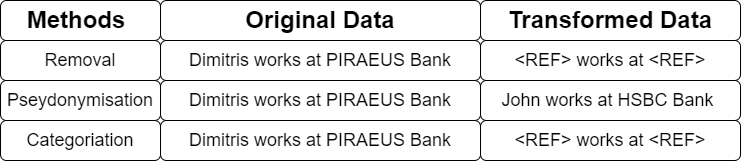}
\caption{Neutralisation Techniques}
\label{Neutralization Techniques}
\end{figure}

%Engaging NER solutions \cite{georgescu2021survey}, such as those grounded in Natural Language Processing (NLP), becomes indispensable in navigating the initial stages of the anonymisation process. NLP techniques present a robust solution, especially given the subjective nature of anonymisation and the scarcity of substantial annotated training data across various domains. Furthermore, NLP facilitates quick adaptability through machine learning, offering tailored solutions on a case-by-case basis, thus enhancing the efficiency and precision of the anonymisation initiative. As we delve deeper into this realm, our current study introduces a transformer model to optimally address the intricacies of the NER task in the context of anonymisation. 

NER methods based on natural Language Processing (NLP) are useful for navigating the initial stages of the anonymisation process \cite{georgescu2021survey}. Considering the nature of anonymisation and the lack of sufficient annotated training data across different domain, NLP methods offer a solid option. Moreover, NLP provides rapid adaptation through machine learning, providing in this way custom solutions and increasing the efficiency and accuracy of the anonymisation process. This study present a transformer model to address the complexity of the NER task in the context of anonymisation.

\subsection{Leveraged Architectures}

%An anonymisation system operates as a pivotal tool in the realm of data privacy, safeguarding sensitive information in various datasets. Fundamentally, it identifies and alters or eradicates personal and confidential data, thereby rendering it inaccessible for unauthorized usage. This system leverages sophisticated algorithms, often rooted in NER and Natural Language Processing (NLP) techniques, to efficiently locate sensitive information such as names, addresses, and financial details. Once identified, the system initiates the neutralization process, where the data undergo one of several treatments: removal, pseudonymisation, or categorization \cite{eke2021pseudonymisation}, \cite{di2021machine}. For the identification several steps are executed. 

Anonymisation services as a crucial tool in the data privacy field, protecting sensitive data in the context of different datasets. Simply, it locates, modifies or removes sensitive information, making the data unusable with out permission. Through anonymisation, sensitive information are located including names, addresses, and more by using algorithms based on NLP approaches. After identification, the data are processed using one of the strategies mentioned above: pseudonymisation, removal and categorization \cite{eke2021pseudonymisation}.
% \cite{di2021machine}. 

\begin{figure}[httb]
\centering
\includegraphics[width=0.97\linewidth]{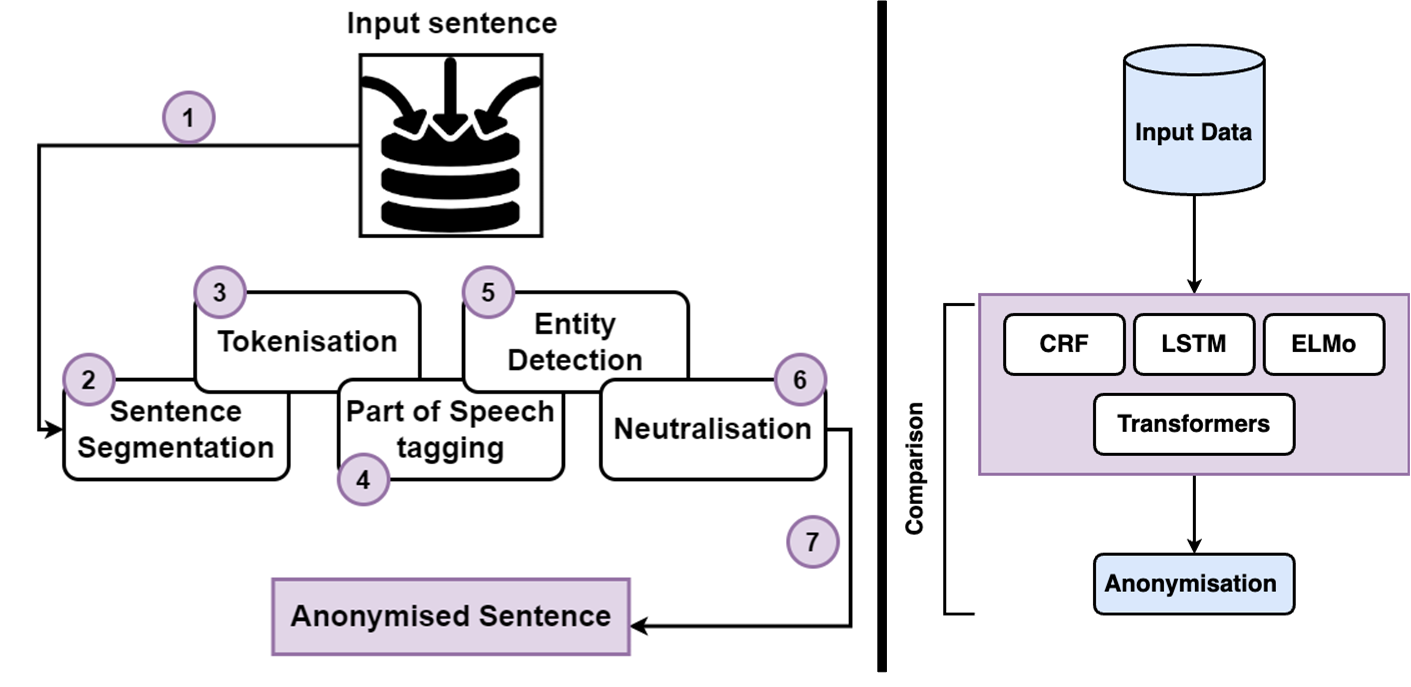}
\caption{System Architecture \& Process}
\label{System_Architecture_and_Process}
\end{figure}

% \begin{figure}[httb]
% \centering
% \includegraphics[width=0.7\linewidth]{Figures/System_architecture.png}
% \caption{System Architecture}
% \label{System Architecture}
% \end{figure}

%Firstly, segmentation is executed to delineate the text into discernible units, often separating sentences and phrases to facilitate focused analysis. Following segmentation, tokenisation takes precedence, where the segmented units are further broken down into individual tokens, typically words or phrases, laying a foundational ground for deeper linguistic scrutiny. Subsequently, part-of-speech tagging is initiated, attributing appropriate grammatical tags to each token, a step that enhances the understanding of the syntactic role each token plays in the sentence structure. Culminating this analytical journey is the entity recognition phase, a critical step where meaningful entities such as names of individuals, organizations, and locations are identified and categorized, thereby extracting salient information and potential insights from the sea of textual data. The end goal is to maintain the utility of the data while ensuring the privacy of individuals. Moreover, it adapts to regulatory requisites, ensuring compliance with data protection laws. Through this sequentially structured process, a detailed and nuanced understanding of text data is achieved, paving the way for diverse applications including, but not limited to, data anonymisation.

In order to enable targeted analysis, segmentation is first utilized to divide the text into pieces. Sentences and phrases are frequently divided differently. The first step is tokenization, where the pieces are further divided into individual tokens, usually words or phrases, making possible in this way the more in-depth analysis. The next phase is part-of-speeching(POS) tagging, where each token takes the proper grammatical tag, improving in this way the comprehension of the syntactic function that it token serves. The entity recognition phase follows, a crucial phase for the anonymisation process, in which the significant entities are recognised and categorised. This process allows the extraction of important data and possible entities from the textual data. The objective is to keep privacy while the data are retrained. Finally, through this process, a detailed understanding of the data is achieved, paving the way for a variety of applications including anonymisation.

\subsection{Solutions}
\label{Solutions}
%Culminating this analytical journey is the entity recognition phase, a critical step where meaningful entities such as names of individuals, organizations, and locations are identified and categorized, thereby extracting salient information and potential insights from the sea of textual data. Building upon these foundational linguistic analyses, many solutions come into play, leveraging sophisticated algorithms, namely, i) CRF, ii) LSTM, iii) ELMo, iv) Transformers, and v) Microsoft Presidio Pre-trained Model, to further enhance text understanding and data processing. 

% The entity recognition phase is a crucial step where significant entities such as names, places, and organisations are recognised and categorised. This step allows the extraction of important data and possible entities from the data. Several solutions are implemented by using advanced algorithms to improve text understanding and data processing. 

The entity recognition phase categorizes important entities like names, places, and organizations, vital for extracting key data, employing advanced algorithms to enhance text comprehension and data processing.

% Four of these solutions

% \begin{itemize}
% \item \textbf{Conditional Random Fields (CRF)}
% \item \textbf{Long Short-Term Memory (LSTM)}
% \item \textbf{Embeddings for Language Model (ELMo) }
% \item \textbf{Transformers }
% \item \textbf{Microsoft Presidio Pre-trained Model}
% \end{itemize}

\subsubsection{CRF}
\label{CRF solution}

Conditional Random Fields (CRF) is a statistical modelling technique that uses a probabilistic framework to accurately predict label sequences given input sample sequences. CRF plays a vital role in applications like text segmentation and NER by facilitating complicated pattern recognition allowing in-depth understanding and classification of complex data patterns.

% \begin{figure}[httb]
% \centering
% \includegraphics[width=0.99\linewidth]{Figures/crf_equation.png}
% \caption{CRF equation}
% \label{CRF equation}
% \end{figure}

\begin{equation}
p(y|x) = \frac{1}{Z(x)}\prod_{t=1}^{T}exp\left \{ \sum_{k=1}^{K}\theta_{k}f_{k}(y_{t},y_{t-1},x_{t}) \right \}
\label{CRF equation}
\end{equation}

\subsubsection{LSTM}
\label{LSTM solution}
%Complementing the capabilities of CRF are LSTM networks, a specialised category within recurrent neural networks designed to handle sequential data with remarkable precision. By capturing temporal dependencies and nuances, LSTM networks can analyse and interpret data points in light of their preceding elements, enhancing the depth of analysis and facilitating applications such as speech and handwriting recognition, where understanding the sequence of data is vital.

LSTM networks are a specialised class of recurrent neural networks that enhance the performance of CRF by handling sequential input with exceptional accuracy. LSTM networks can analyse and interpret data points in light of their preceding elements by capturing temporal dependencies and nuances. This improves the depth of analysis and makes applications like speech and handwriting recognition easier, as comprehending the sequence of data is crucial.

\subsubsection{ELMo}
\label{ELMo solution}
%Adding another layer of sophistication is ELMo, a method that generates deep contextualized word representations, going beyond traditional word embedding techniques to offer rich semantic representations grounded in a deep understanding of words in varying contexts. ELMo appreciates the dynamic nature of language, capturing the nuances that come with different contextual uses of words, thus paving the way for more precise and semantically rich text analysis.

ELMo , a beyond conventional word embedding technique used to build deep contextualised word representations, delves more in the area of text anonymisation. It provides representations based on in-depth understanding of words in various scenarios. This model recognises the dynamic language and captures the nuances that occur making for efficient and accurate text analysis.

\subsubsection{Transformers}
\label{Solution using transformers}

%The technical trajectory logically leads us to the introduction of transformer models, a further step in the field of natural language processing, in comparison with CRF, LSTM and ELMo. While the solutions laid a strong foundation, offering robust mechanisms for pattern recognition, sequential data handling, and deep contextualized word representations, transformer models revolutionise the way language data are processed. Transformers brought the attention mechanism, emphasising the relationships between different words in a sentence, thereby facilitating a deeper understanding of contextual nuances. Furthermore, transformers eliminate the necessity of sequential data processing, enabling parallel computation and drastically reducing training times. This architectural advancement combines the strengths of its predecessors and further enriches it with self-attention mechanisms and deep learning layers. 

Unlike the rest the technical development brings us to transformer models which are an advancement in the field of natural language processing. Though the solutions delivered appropriate mechanisms for deep contextualised word representation, sequential data handling and pattern recognition, transformer models redefine the semantic processing. They offered a great solution, emphasizing the connections between the words in sentences and enabling more nuanced understanding within diverse contexts. They also eliminate the sequential processing of the data, which enables parallel computations and greatly shortens training time. With the self-attention mechanisms and deep learning layers, this architectural novelty becomes much more powerful; it greatly increases the performance of its predecessors.

\subsection{Transformers Architecture}
\label{Transformers Architecture}

%Transformers \cite{khan2022transformers} are a type of deep learning model that have revolutionised natural language processing tasks, including the critical domain of textual and numerical anonymisation. At the heart of transformers lies the attention mechanism, which can weigh the influence of different words on each other in a given text, allowing the model to capture intricate patterns and dependencies in the data. This capability is leveraged in the context of anonymisation to identify and replace sensitive information, such as names, addresses, and financial details, with generic placeholders, thereby protecting individual privacy. 

The field of NLP tasks, particularly textual and numerical anonymisation, has been improved by deep learning models known as transformers \cite{khan2022transformers}. In order for the model to identify complex patterns and relationships in the data, the attention mechanism  is able to evaluate the impact that various words have on one another inside a specific text. Identifiable and generic placeholders are used in place of sensitive data,  protecting the privacy of the individual. 

\begin{figure}[httb]
\centering
\includegraphics[width=0.5\linewidth]{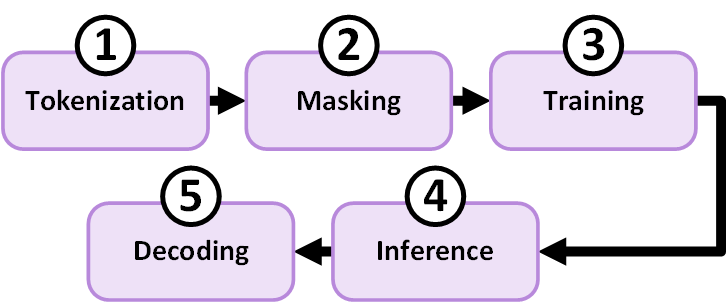}
\caption{Transformers technique step-by-step}
\label{Transformers technique}
\end{figure}

%Numerical anonymisation, on the other hand, involves the masking or perturbation of numerical data to prevent the revelation of individual identities or sensitive attributes. Transformers can be trained to recognise contexts where numerical data appears and apply sophisticated anonymisation techniques, such as k-anonymity or differential privacy, to maintain the utility of the data while ensuring privacy. The self-attention layers in the transformer architecture facilitate the understanding of complex patterns and relationships in both textual and numerical data, making transformers a powerful tool in the creation of anonymised datasets that balance the need for privacy with the preservation of data utility.

In contrast, numerical anonymisation refers to the hiding or distortion of numbers in order not to leak any personal identification information or confidential traits. Transformers can be easily taught to detect many entities that contain numerical data and then employ various sophisticated anonymisation techniques such as differential privacy or k-anonymity to protect the security of the information without impacting its practicality. However, transformers are a very useful means of generating anonymised datasets that meet in the middle ground between privacy and data effectiveness due to self-attention layers in the transformer architecture.

\section{Experimental Results}
\label{Experimental Results}
This section delineates the evaluation results and discussion for the implemented methods, providing an in-depth comparative analysis. 
% In this research, we delineate a comparative analysis of four prominent machine learning models, namely, CRF, LSTM, ELMo, and Transformers, utilising them to carry out textual and numerical anonymisation tasks. Initially, the datasets will be meticulously partitioned into training, validation, and testing subsets to facilitate a robust evaluation. Each model will undergo a stringent hyperparameter tuning process, optimized based on their performance on the validation set. Following this, we will engage in the training of each model utilising the respective training subsets, and consequent evaluation on the test subsets to acquire critical metrics.

% \begin{figure}[httb]
% \centering
% \includegraphics[width=0.4\linewidth]{Figures/paper_models.png}
% \caption{Methodology Process}
% \label{Methodology Process}
% \end{figure}

\subsection{Metrics}
\label{Metrics}

This work leverages quantitative metrics in order to concisely evaluate the results of the utilised models on the task of anonymisation. In particular, the i) precision, ii) recall, and iii) F1-score metrics are adopted to provide this evaluation. This methodology intends to provide a comparison while also illuminating the best ways to use machine learning models for textual and numerical anonymisation

% These metrics serve as the benchmark for our comparative analysis, assisting in drawing a comprehensive understanding of each model's efficacy in the realm of anonymisation. This methodology intends to provide a comparison while also illuminating the best ways to use machine learning models for textual and numerical anonymisation. This will pave the way for the development of more secure and effective data handling procedures. 

The Recall metric (Eq.~\ref{recall}), measures the proportion of actual positives that are correctly identified as positive by the model.

\begin{equation}
\label{recall}
Recall = \frac{TP}{TP + FN}
\end{equation}

Where TP is True Positives, TN is True Negatives, FP is False Positives and FN is False Negative. Precision (Eq.~\ref{precision}) measures the correctly predicted instances out of all instances predicted as positive. 
% It quantifies the model's ability to avoid false positives.

\begin{equation} 
\label{precision}
 Precision = \frac{TP}{TP + FP}
\end{equation}

The F1 score (Eq.~\ref{f1}) captures the balance between true positive rate (TPR) and precision. 

% Precision is defined as the ratio of true positives to the sum of true positives and false positives.

\begin{equation} 
\label{f1}
 F1 = \frac{2 \times TP}{2 \times TP + FP + FN}
\end{equation}

% \section{Results}
% \label{Results}

\subsection{Dataset}
\label{Dataset}

% \begin{figure}[httb]
% \centering
% \includegraphics[width=0.99\linewidth]{Figures/data_preview.png}
% \caption{Dataset Preview}
% \label{Dataset}
% \end{figure}

\begin{table}[]
\centering
\caption{Dataset Sample}
\begin{tabular}{lllll}
\rowcolor[HTML]{EFEFEF} 
\textbf{} & \textbf{Sentence \#} & \textbf{Word} & \textbf{POS} & \textbf{Tag} \\ \hline
0         & Sentence 1          & Thousands     & NNS          & O            \\
1         & Sentence 1          & of            & IN           & O            \\
2         & Sentence 1          & demonstrators  & NNS          & O            \\
3         & Sentence 1          & have          & VBP          & O            \\
4         & Sentence 1          & marched       & VBN          & O            \\
5         & Sentence 1          & through       & IN           & O            \\
6         & Sentence 1          & London        & NNP          & B-geo        \\
7         & Sentence 1          & to            & TO           & O            \\
8         & Sentence 1          & protest       & VB           & O            \\
9         & Sentence 1          & the           & DT           & O            \\
10        & Sentence 1          & war           & NN           & O            \\
11        & Sentence 1          & in            & IN           & O            \\
12        & Sentence 1          & Iraq          & NNP          & B-geo       
\end{tabular}
\label{Dataset Sample}
\end{table}

%The dataset, that was used for the experiments, at hand encompasses a meticulous breakdown of sentences tailored for NER applications. Each record in the dataset corresponds to a distinct word, accompanied by its respective Part-of-Speech (POS) tag and NER annotation. The column labelled 'Sentence \#' demarcates the commencement of each new sentence, with subsequent words in the same sentence not repeating this identifier. This structured format ensures clarity in the sequential arrangement of words within their contextual sentences. By offering a comprehensive amalgamation of words, their grammatical classifications, and their NER designations, this dataset stands as a pivotal resource for the training, evaluation, and refinement of NER-based machine learning models. Its composition and granularity make it an indispensable tool for advancing research and applications in the realm of natural language processing.

The dataset used for the experiments in question includes carefully formulated sentences designed specifically with NER applications. The dataset comprises of a record for every separate word with its POS tag and also NER annotation. The ‘Sentence \#' column indicates the beginning of a sentence and subsequent words in the same sentences do not replicate this marker. A sequential order of sentences through the words is ensured by the structured structure. With a complex concoction of words, their classifications in terms of grammar and the NER marks put on them this dataset proves to be one remarkable resource for training, assessing as well as fine-tuning. With its composition and granularity, it is indeed an inevitable tool in the advancement of research and applications for use within the field of natural language processing.

% \begin{figure}[httb]
% \centering
% \includegraphics[width=0.8\linewidth]{Figures/tags.png}
% \caption{Tags used in the dataset}
% \label{Tags}
% \end{figure}

Table \ref{Tag} provides a detailed breakdown of the tags utilized within the dataset for NER tasks. Each tag signifies a specific type of named entity or its position within such an entity. 

\begin{table}[htbp]
    \centering
    \caption{Tag Descriptions}
    \label{Tag}
    \begin{tabular}{|c|p{5cm}|}
        \hline
        \textbf{Tag} & \textbf{Description} \\
        \hline
        O & Other \\
        \hline
        B-geo & Beginning of Geographical Entity \\
        \hline
        B-gpe & Beginning of Geopolitical Entity \\
        \hline
        B-per & Beginning of Person \\
        \hline
        I-geo & Inside Geographical Entity \\
        \hline
        B-org & Beginning of Organisation \\
        \hline
        I-org & Inside Organisation \\
        \hline
        B-tim & Beginning of Time \\
        \hline
        B-art & Beginning of Artifact \\
        \hline
        I-art & Inside Artifact \\
        \hline
        I-per & Inside Person \\
        \hline
        I-gpe & Inside Geopolitical Entity \\
        \hline
        I-tim & Inside Time \\
        \hline
        B-nat & Beginning of Natural Phenomenon \\
        \hline
        % B-eve & Beginning of Event \\
        % \hline
        % I-eve & Inside Event \\
        % \hline
        % I-nat & Inside Natural Phenomenon \\
        % \hline
    \end{tabular}
\end{table}

For instance, tags prefixed with 'B-' indicate the beginning of a named entity, while those prefixed with 'I-' denote words inside a named entity. The tags range from generic classifications, such as 'O' for words not associated with any special entity, to more specific categories like geographical entities, geopolitical entities, organizations, and time expressions, among others. Such a granular tagging system allows for a nuanced understanding of the contextual relevance of words within the dataset. This table serves as a foundational reference for interpreting the NER annotations and is crucial for any researcher or practitioner working with the dataset.

\subsection{Experimental Results for CRF, LSTM and ELMo}

In the experiment conducted to compare the efficacy of different machine learning models in the realm of textual and numerical anonymisation, distinct variations in performance were observed. The CRF model showed excellent performance in segmenting and labelling the sequence data, getting better results while anonymizing. The second best was the LSTM model that trained efficiently long-term dependencies in the sequence data, but underperformed slightly to CRF. ELMo model, on the other hand was lagging behind with scores of 0.74, 0.81 and 0,77. representing underperformance in terms of detecting intricacies of the data as compared to other models. However, due to its deeply contextualized word representation it did not meet the strict requirements of anonymization missions. The two leading models were the CRF and the transformers, achieving the highest precision recall value and F1-scores. The LSTM model also showed a big capability in capturing the long-term dependencies of the data with scores of 0.93 for precision and recall, while getting an F1 score of 0 The lowest results were with the ELMo model which achieved 0.74, 0.81 and 0.77 as scores showing a relative weakness to meet clearly demanding pre-requisites of anonymization tasks.

\subsection{Experimental Results for Transformers}
\label{Experimental Results using Transformers}

% \begin{figure}[httb]
% \centering
% \includegraphics[width=0.7\linewidth]{Figures/Transformers_metrics.png}
% \caption{Evaluation Metrics using Transformers}
% \label{Transformers metrics}
% \end{figure}

%In the meticulous evaluation of the transformer model in the context of textual and numerical anonymisation tasks, we observed a commendable performance characterized by a balanced proficiency in both precision and recall. Garnering a weighted average precision score of 0.93 demonstrates the model's adeptness at accurately identifying and handling sensitive textual and numerical data, minimising false positive identifications. Its recall score of 0.93  reflects a robust capability to retrieve a substantial portion of the relevant entities present in the dataset. This is complemented by an F1-score of 0.93 as well, which stands as a testament to the model's harmonized performance in precision and recall, ensuring a reliable anonymisation process. The self-attention mechanism intrinsic to the transformer architecture aids in discerning complex patterns and relationships in the data, thereby facilitating this balanced performance. While there is room for optimization to reach the pinnacle of efficiency exhibited by some other models, the transformer’s present scores underline its potent utility as a reliable tool in the data anonymisation landscape, promising robust protection of sensitive information while maintaining a high degree of data utility.

The intricate analysis of the transformer model within the framework of textual and numerical anonymisation tasks demonstrated an impressive performance with a profound accuracy in precision-recall terms. Obtaining a weighted average precision score of 0.93, this model is able to efficiently discover and encode sensitive textual data as well as the numerical ones with minimum false positive detects It has a recall score of 0.93, thus demonstrating its ability to receive the majority of relevant entities in the dataset. This is further reinforced by an F1-score of 0.93, which demonstrates the model’s coordinated effectiveness in precision and also recall that guarantee a consistent deidentification procedure. This balanced performance is achieved in part by the self-attention mechanism inherent to the transformer architecture, which helps uncover intricate patterns and relationships within the given data. Although there may be some opportunities for improvement in order to reach the peak of proficiency demonstrated by some other models, the current results that are pioneered indicate high performance capabilities which make this model a very valuable asset among data anonymisation tools delivering a reasonable protection level while also preserving a significant value and quality.

\subsection{Comparison between Models}
\label{Comparison between Models}

\begin{figure}[httb]
\centering
\includegraphics[width=0.9\linewidth]{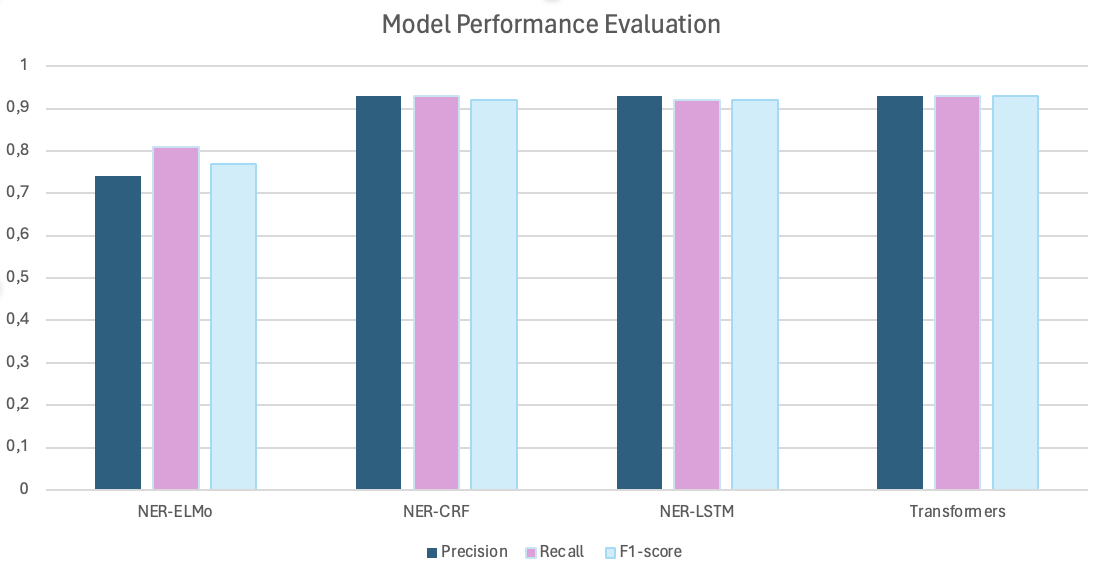}
\caption{Evaluation Metrics for all models}
\label{All metrics}
\end{figure}

%In our comparative evaluation across various NER models, distinct performance patterns were observed. The NER-CRF model and the transformers exhibited superior performance, achieving the highest scores across all metrics: Precision 0.93, Recall 0.93, and F1-score 0.93. NER-LSTM closely followed, showcasing robust results with scores of 0.93, 0.92, and 0.92 for Precision, Recall, and F1-score, respectively. 
The performance patterns displayed across several NER models are very different in our comparative analysis. The NER-CRF model and the transformers exhibited superior performance, achieving the highest scores across all metrics: F-measure is 0.93, Precision: 0.93 and Recall of 0.93. NER-LSTM was closely approached by, demonstrating excellent results with a 0.93, 0.92 and 0.92 in Precision, Recall and F1 score, respectively. The NER-ELMo model demonstrated commendable outcomes, particularly in Recall 0.81, though its Precision and F1-score were slightly lower at 0.74 and 0.77. This comparison underscores the nuanced capabilities of each model, highlighting the trade-offs between Precision, Recall, and F1-score, thereby providing valuable insights for model selection based on specific NER tasks and also the evolution of the transformers in the field of anonymisation.

\begin{table}[http]
\centering
\caption{Evaluation metrics for all models}
\label{tab:ner_metrics}
\begin{tabular}{|c|c|c|c|}
\hline
\textbf{} & \textbf{Precision} & \textbf{Recall} & \textbf{F1-score} \\
\hline
\textbf{NER-ELMo} & 0.74 & 0.81 & 0.77 \\
\hline
\textbf{NER-CRF} & \textbf{0.93} & \textbf{0.93} & 0.92 \\
\hline
\textbf{NER-LSTM} & \textbf{0.93} & 0.92 & 0.92 \\
\hline
\textbf{Transformers} & \textbf{0.93} & \textbf{0.93} & \textbf{0.93} \\
\hline
\end{tabular}
\end{table}

% In the evaluation of text anonymisation methods, as presented in Table \ref{tab:ner_metrics}, the NER-CRF model and the Transformers emerged as the most proficient, achieving the highest scores across all metrics. While NER-LSTM also showcased commendable performances, NER-ELMo trailed slightly behind in terms of precision and F1-score. These findings emphasise the varied efficacies of the models in addressing NER challenges.

\section{Conclusion}
\label{Conclusion}
As the digital landscape continues to evolve, the importance of effective text anonymisation cannot be understated. Our research went deep into various models, with a spotlight on CRF, LSTM, ELMo, and Transformers, to discern their capacities to preserve data privacy without compromising the richness of textual data. The comparative analysis underscored the prowess of the CRF model but also has shown significant promise in certain contexts using Transformers. While each model brings its unique strengths to the table, the variability in performance emphasizes the need for continued research in this domain. As data privacy concerns rise in prominence, this study is highlighting both the challenges and potential solutions in the realm of text anonymisation. 

Future endeavours should not only build upon these findings but also explore the integration of these models to harness their combined strengths, paving the way for more robust and efficient text anonymisation techniques.

\section*{Acknowledgment}
This project has received funding from the European Union’s Horizon Europe research and innovation programme under grant agreement No. 101070181 (TALON).

\bibliographystyle{IEEEtran} % IEEE style
\bibliography{refs}

\end{document}